**ARTICLE** OPEN

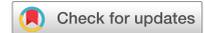

# Integrated multimodal artificial intelligence framework for healthcare applications

Luis R. Soenksen [1,2,5], Yu Ma [3,5], Cynthia Zeng [3,5], Leonard Boussioux [3,5], Kimberly Villalobos Carballo [3,5], Liangyuan Na [3,5], Holly M. Wiberg [3], Michael L. Li [3], Ignacio Fuentes [1] and Dimitris Bertsimas [1,3,4✉]

Artificial intelligence (AI) systems hold great promise to improve healthcare over the next decades. Specifically, AI systems leveraging multiple data sources and input modalities are poised to become a viable method to deliver more accurate results and deployable pipelines across a wide range of applications. In this work, we propose and evaluate a unified Holistic AI in Medicine (HAIM) framework to facilitate the generation and testing of AI systems that leverage multimodal inputs. Our approach uses generalizable data pre-processing and machine learning modeling stages that can be readily adapted for research and deployment in healthcare environments. We evaluate our HAIM framework by training and characterizing 14,324 independent models based on HAIM-MIMIC-MM, a multimodal clinical database ($N = 34,537$ samples) containing 7279 unique hospitalizations and 6485 patients, spanning all possible input combinations of 4 data modalities (i.e., tabular, time-series, text, and images), 11 unique data sources and 12 predictive tasks. We show that this framework can consistently and robustly produce models that outperform similar single-source approaches across various healthcare demonstrations (by 6–33%), including 10 distinct chest pathology diagnoses, along with length-of-stay and 48 h mortality predictions. We also quantify the contribution of each modality and data source using Shapley values, which demonstrates the heterogeneity in data modality importance and the necessity of multimodal inputs across different healthcare-relevant tasks. The generalizable properties and flexibility of our Holistic AI in Medicine (HAIM) framework could offer a promising pathway for future multimodal predictive systems in clinical and operational healthcare settings.

## INTRODUCTION

Artificial intelligence (AI) and machine learning (ML) systems are poised to become fundamental tools in next-generation clinical practice and healthcare operations[1]. Such anticipated utility, particularly in AI/ML systems aimed to improve clinical efficiency and patient outcomes, will require knowledge from multiple data sources and various input modalities[2–4]. Multimodal architectures for AI/ML systems are attractive because they can emulate the input conditions that clinicians and healthcare administrators currently use to perform predictions and respond to their complex decision-making landscape[2,5]. A typical clinical practice uses a diverse set of information formats contained within the patient electronic health record (EHR) such as tabular data (e.g., age, demographics, procedures, history, billing codes), image data (e.g., photographs, x-rays, computerized-tomography scans, magnetic resonance imaging, pathology slides), time-series data (e.g., intermittent pulse oximetry, blood chemistry, respiratory analysis, electrocardiograms, ultra-sounds, in-vitro tests, wearable sensors), structured sequence data (e.g., genomics, proteomics, metabolomics) and unstructured sequence data (e.g., notes, forms, written reports, voice recordings, video) among other sources[6]. Recently, AI/ML models leveraging multiple data modalities have been demonstrated for the domains of cardiology[7–9], dermatology[10], gastroenterology[11], gynecology[12], hematology[13], immunology[14], nephrology[15], neurology[16,17], oncology[18–20], ophthalmology[21], psychiatry[22], radiology[23–25], public health[26] and healthcare operational analytics (i.e., mortality, length-of-stay, and discharge predictions)[27–30]. Furthermore, it has been shown that multimodality in most of these domains can increase the performance of AI/ML systems (accuracy: 1.2–27.7%) compared to single-modality approaches for the same task[2]. However, developing unified and scalable pipelines that can consistently be applied to train multimodal AI/ML systems that leverage and outperform their single-modality counterparts has remained challenging[2]. This motivates the development of our Holistic Artificial Intelligence in Medicine (HAIM) framework, a modular ML pipeline (Fig.1) that can be adapted to receive standard EHR information from multiple input data modalities (i.e., tabular data, images, time-series, and text). Our HAIM framework addressed the need for a more generalizable methodology to create this class of systems. It can leverage user-defined pre-trained feature-extraction models as part of a unified processing and feature aggregation stage that allows for simple and scalable downstream modeling of a variety of clinically relevant predictive tasks. Based on this pipeline, we build and test thousands of classification models with sample EHR inputs to systematically investigate the value of adding individual data modalities to these systems. To our knowledge, this has not yet been analyzed to greater detail in prior clinical multimodal AI/ML demonstrations. We provide this work as an open-source codebase for clinicians and researchers in the hope it will allow them to train and test AI/ML systems more easily with the local datasets, pre-trained feature extractors, and clinical questions of their choosing to fully leverage multimodality at their institutions.

[1]Abdul Latif Jameel Clinic for Machine Learning in Health, MIT, Cambridge, MA 02139, USA. [2]Wyss Institute for Biologically Inspired Engineering, Harvard University, Boston, MA 02115, USA. [3]Operations Research Center, Massachusetts Institute of Technology (MIT), Cambridge, MA 02139, USA. [4]Sloan School of Management, MIT, Cambridge, MA 02139, USA. [5]These authors contributed equally: Luis R. Soenksen, Yu Ma, Cynthia Zeng, Leonard Boussioux, Kimberly Villalobos Carballo, Liangyuan Na. ✉email: dbertsim@mit.edu

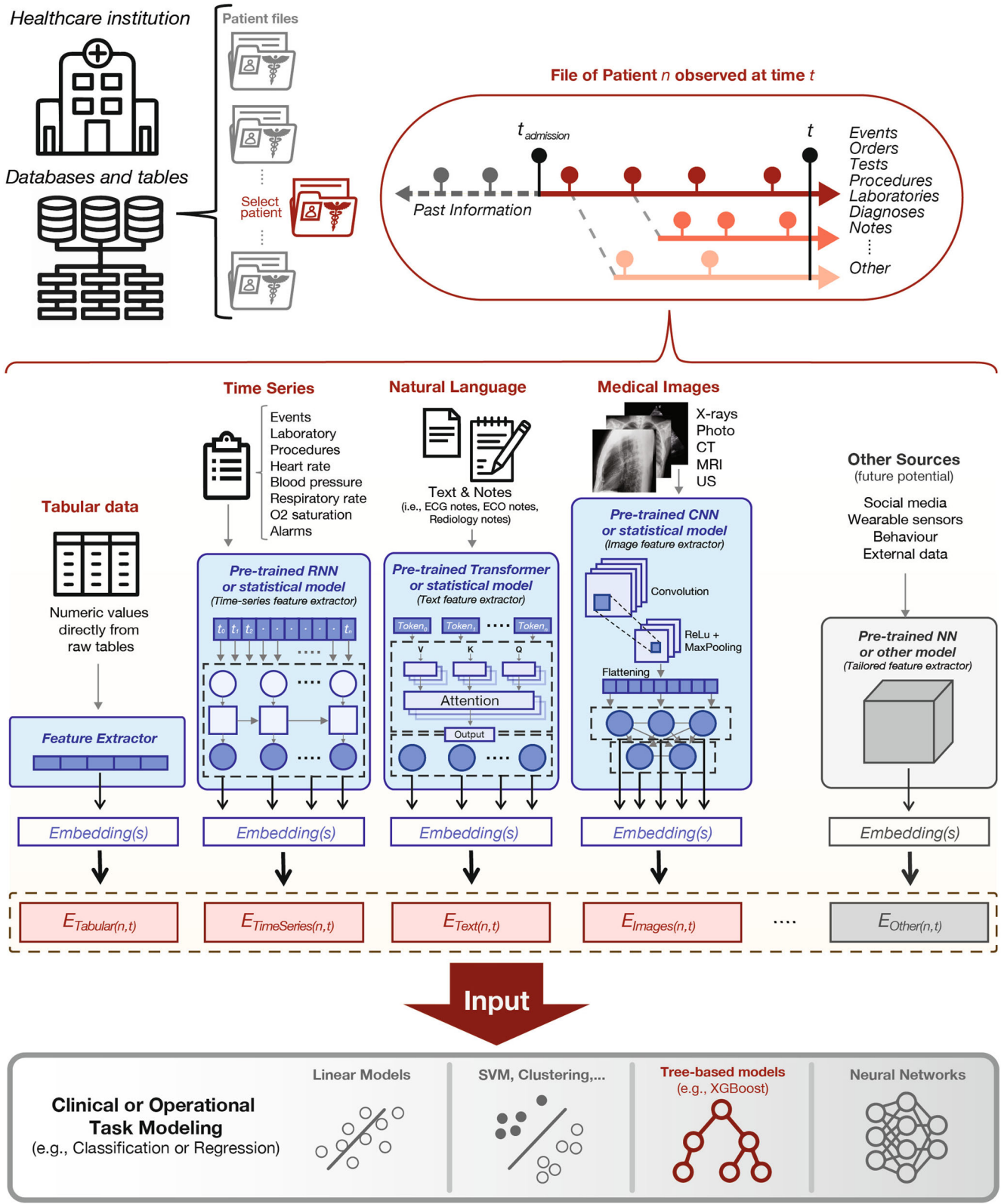

## RESULTS

### Demonstration of HAIM framework on multimodal clinical dataset

We demonstrate the feasibility and versatility of the HAIM framework on a compiled multimodal dataset (HAIM-MIMIC-MM),

which includes a total of 34,537 samples involving 7279 hospitalization stays and 6485 unique patients. We summarize the general characteristics of HAIM-MIMIC-MM (i.e., number of samples and features) in Table 1. Qualitatively, our HAIM framework appears to improve on previous work in this field[30] by

**Fig. 1 Holistic Artificial Intelligence in Medicine (HAIM) framework.** Under this framework, databases and tables sourced from specific healthcare institutions such as HAIM-MIMIC-MM combined from MIMIC-IV and MIMIC-CXR-JPG for this work are processed to generate individual patient files. These files contain past and present multimodal patient information from the moment of admission. For processing under the HAIM framework, every data modality is fed to independent embedding generating streams. In this work, tabular data is minimally processed using simple transformations or normalizations to produce encodings or embedding-like categorical numerical values ($E_{Tabular(n,t)}$, where $n$ = unique stay/hospitalization/patient and $t$ = sampling time). Selected time-series are processed by generating statistical metrics on each of the time-dependent signals to produce embeddings representative of their trends ($E_{TimeSeries(n,t)}$) from the moment of admission until the sampling time. Natural language inputs such as notes are processed using a pre-trained transformer neural network to generate text embeddings of fixed size ($E_{Text(n,t)}$). Furthermore, image inputs such as X-rays are processed using a pre-trained convolutional neural network to also extract fixed-size embeddings out of the model output probability vectors and dense features ($E_{Images(n,t)}$). While not done in this work, thanks to the modularity of the embedding extraction process in the HAIM framework, other pre-trained models or systems could be added to generate embeddings from other types of data sources if needed ($E_{Other(n,t)}$). All generated embeddings are concatenated to generate a fusion embedding, which can be used to train, test, and deploy models for predictive analytics in healthcare operations. For this work, we tested and utilized only XGBoost as a canonical type of architecture for building the downstream predictive models based on fusion embeddings. CNN Convolutional Neural Network, CT Computerized Tomography, ECG Electrocardiogram, ECO Echocardiogram, MRI Magnetic Resonance Imaging, NN Neural Network, O2 Oxygen, ReLU Rectified Linear Unit, RNN Recurrent Neural Network, US Ultrasound.

**Table 1.** General characteristics of the HAIM-MIMIC-MM database.

| Characteristic | MIMIC-IV-MM |
|---|---|
| # Samples | 34537 |
| # Demographic Variables | 6 |
| # Chart Event Variables | 9 |
| # Laboratory Event Variables | 23 |
| # Procedure Event Variables | 10 |
| # X-ray Variables | 1 |
| # Text Note Variables | 3 |

HAIM-MIMIC-MM is a combination of MIMIC-IV and MIMIC Chest X-ray filtered to only include patients that have at least one chest X-ray performed with the goal of validating multimodal predictive analytics in healthcare operations. The number of samples and quantities of variables are described. Demographic features correspond only to a tabular data modality, while chart, laboratory, and procedure events correspond to time-series. X-ray variables correspond to types of medical images, while text note variables correspond to the test in radiology, electrocardiogram, and echocardiogram natural language reports.

including scalable patient-centric data pre-processing and enabling standardized feature extraction stages that allow for rapid prototyping, testing, and deployment of predictive models based on user-defined prediction targets. Our HAIM framework displays consistent improvement on average AUROC (Fig. 2a color gradient) across all models as the number of modalities and data sources increases. Furthermore, the trend of reducing AUROC standard deviation (SD) values also appears to follow from increasing the number of modalities and data sources (Fig. 2a greyscale gradient). We also report Receiver Operating Characteristic (ROC) curves for the best found single-modality predictive models (Fig. 2c) as compared with typical multimodal predictive models based on the HAIM framework (Fig. 2b). All 14,324 individual model AUROCs (10,230 for chest diagnosis prediction tasks, 2047 for length-of-stay and 2047 mortality prediction) are shown along with their respective SDs in Supplementary Fig. 1A–D. These results suggest that our HAIM framework can consistently improve predictive analytics for various applications in healthcare as compared with single-modality analytics. Quantitatively, Fig. 3a, b shows that our HAIM framework produces models with multisource and multimodality input combinations that improve from average performance of canonical single-source (and by extension single-modality) systems for chest x-ray pathology prediction ($\Delta_{AUROC}$: 6–22%), length-of-stay ($\Delta_{AUROC}$: 8–20%) and 48 h mortality ($\Delta_{AUROC}$: 11–33%). Specifically, for chest pathology prediction, the minimum per task improvements include: Fracture ($\Delta_{AUROC} = 6\%$), Lung Lesion ($\Delta_{AUROC} = 7\%$), Enlarged Cardio mediastinum

($\Delta_{AUROC} = 9\%$), Consolidation ($\Delta_{AUROC} = 10\%$), Pneumonia ($\Delta_{AUROC} = 8\%$), Atelectasis ($\Delta_{AUROC} = 6\%$), Lung Opacity ($\Delta_{AUROC} = 7\%$), Pneumothorax ($\Delta_{AUROC} = 8\%$), Edema ($\Delta_{AUROC} = 10\%$) and Cardiomegaly ($\Delta_{AUROC} = 10\%$). Furthermore, the average percent improvement of all multimodal HAIM predictive systems is 9–28% across all evaluated tasks (Fig. 3a). All AUROC-related results displayed in Figs. 2a and 3a, b are grouped and ordered by number of modalities (range = 1–4, encompassing tabular, time-series, text, and images), number of data sources (range = 1–11, including each individual data source in HAIM-MIMIC-MM) and sample size (N) for ease of analysis.

**Analysis of source and multimodality contributions on model performances**

To understand how each data source and modality contributes to the final performance, we calculate Shapley values[31] of each of the 11 sources and 4 modalities as it contributes to the final AUROC test-set performance. Since our demonstrated predictive tasks are treated as binary classification problems, we assumed that the AUROC of a model with no data source is 0.5, and the AUROC of the model of a particular modality is the average AUROC of the models of all sources that belong to such modality. Aggregated Shapley values for all data modalities per predictive task are reported in Fig. 3c, while Shapley values for all data sources per predictive task are shown in Supplementary Fig. 2. Different tasks exhibit distinct distributions of aggregated Shapley values across data modalities and sources. In particular, we observe that vision data contributed most to the model performance for the chest pathology diagnosis tasks, but for predicting length-of-stay and 48 h mortality, the patient's historical time-series records appeared to be the most relevant. Shapley values also provide a way to monitor errors and information loss propagation during the feature extraction and model training phases of our HAIM framework. Data modalities associated with small (or negative) Shapley values indicate either an absence of extracted information or error propagation leading to detrimental local effects on downstream model performance (Fig. 3b and Supplementary Fig. 2). This situation can be potentially addressed by removing such input data modalities or by selecting different pre-trained feature extraction models specific to that data modality. Nevertheless, we see that across all tasks, in our specific sample HAIM-MIMIC-MM demonstrations, every single-modality contributes positively to a monotonic trend with diminishing returns on the predictive capacity of the models (Fig. 3a and c), likely due to multimodal data redundancy. These observations attest to the potential value (and limitations) of using multimodal inputs and pre-trained feature extraction modules in frameworks like HAIM, which could be used to generate predictive models for diverse clinical tasks more cost-effectively than previous strategies. A high-level schematic of the complete HAIM pipeline for training and evaluation of models throughout this work is described in Fig. 3d. The general process of HAIM-MIMIC-MM database preparation, as well as embedding extraction and fusion that serves as input for this pipeline, can be found in Fig. 1.

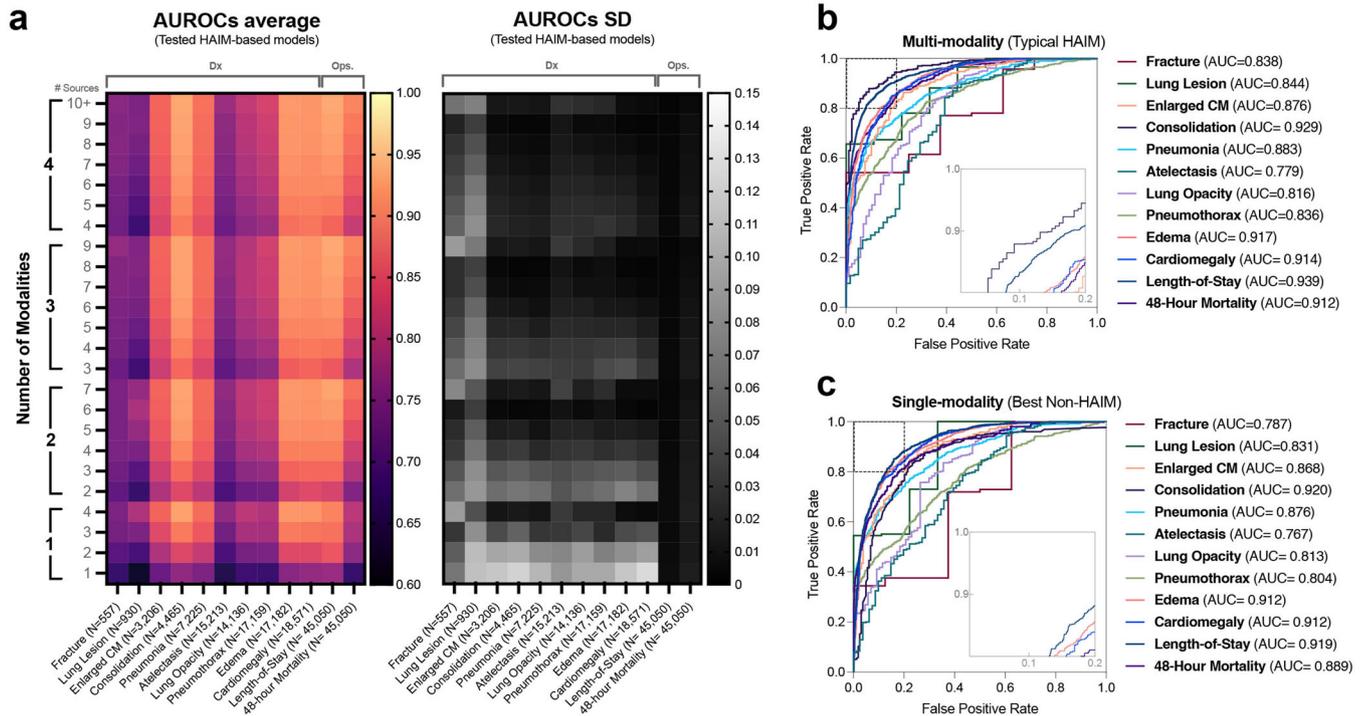

**Fig. 2 Performance of the multimodal HAIM framework on various demonstrations for healthcare operations. a** Average and standard deviation values of the area under the receiver operating characteristic (AUROC) for all demonstrations including pathology diagnosis (i.e., lung lesions, fractures, atelectasis, lung opacities, pneumothorax, enlarged cardio mediastinum, cardiomegaly, pneumonia, consolidation, and edema), as well as length-of-stay and 48 h mortality prediction. The number of modalities refers to the coverage among tabular, time-series, text, and image data. The number of sources refers to the coverage among available input data sources (10 for pathology diagnosis, while 11 for length-of-stay and 48 h mortality prediction). Thus, the position (Modality = 2, Sources = 3) corresponds to the average AUROC of all models across all input combinations covering any 2 modalities using any 3 input sources. Increasing gradients on average AUROC appear to follow from increasing the number of modalities and number of sources across all evaluated tasks. Decreasing gradients on AUROC standard deviations follow from less variability in performance as a higher number of modalities and data sources is used. **b** Receiver operating characteristic (ROC) curves for typical HAIM model across all use cases exhibiting input multimodal. **c** ROC curves for a best-performing model with single-modality inputs across the same use cases. Consistent averaged improvements across all tasks are observed in multimodality as compared to single-modality systems. AUROC Area under the curve, AUROC Area under the receiver operating characteristic curve, CM Cardiomediastinum. Dx Diagnosis, HAIM Holistic Artificial Intelligence in Medicine, Ops Operations, SD Standard deviation.

## DISCUSSION

Inferring latent features from rich and heterogeneous multimodal EHR information could provide clinicians, administrators, and researchers with unprecedented opportunities to develop better pathology detection systems, actionable healthcare analytics, and recommendation engines for precision medicine. Our results directly illustrated that different data modes are more useful for different tasks, and thus a multimodal approach to construct a comprehensive pipeline for AI/ML in healthcare. In addition to leveraging multimodal inputs, our HAIM framework attempts to solve several bottleneck challenges in this kind of AI/ML pipeline for healthcare in a more unified and robust way than previous implementations, including the possibility of working with tabular and non-tabular data of unknown sparsity from multiple standardized and unstandardized heterogeneous data formats. The use of fusion embeddings obtained directly from individual patient files suggests that a HAIM framework can potentially facilitate the definition, testing, and deployment of AI/ML models that may be useful for managing complex clinical situations and day-to-day practice in healthcare systems. More specifically, if implemented across many predictive tasks while using the same patient embeddings, this approach could potentially help accelerate the advent of scalable predictive systems to improve patient outcomes and quality of care. From these observations, our work distinguishes itself from previously published systems in three main ways: (A) First, our work systematically investigates the value of progressively adding data modalities and sources to clinical multimodal AI/ML systems in much greater detail and larger combinatorial input space than any prior investigation of such class systems. Previous works in this field assume advantageous properties to multimodality without clear validation of the dynamics of such expected performance benefits as data modalities are added. Through our investigation by conducting 14,324 model experiments with different input modalities and data source combinations, we provide strong empirical evidence that supports the potential for reaching such positive monotonic trends in performance from multimodal AI/ML systems as data modalities are added. However, our investigation also unveils previously unreported local non-monotonic and diminishing return effects on the predictive capacity of these models under certain conditions of data source availability, error, and redundancy, which are relevant and can become interpretable through our use of aggregated Shapley values during analysis. (B) Second, our data pre-processing and modeling pipeline expands on the notion of high modularity from previously published work, that tend to employ ad-hoc multimodal architectures trained directly on fused data inputs, which are usually closed, less compatible with other datasets, and modeling changes across users. Instead, our approach leverages externally validated open-sourced models as feature extractors to create unified vector representations of patient files that allow for much simpler downstream modeling of target variables. Furthermore, this framework enables and encourages users to update selected feature extractors more easily with new state-of-the-art (SOTA) or more advantageous methods as the community develops them, without requiring to re-train other feature extractors. (C) Finally, our work demonstrates one of the highest numbers of sources and data modalities used so far in multimodal clinical AI/ML systems for EHRs, including tabular data, time-series, text, and images along with the use of interpretability techniques such as Shapley values. Using aggregated Shapley values, we can quantitatively establish the importance and heterogeneity of different data sources and modalities across a large number of experiments in different healthcare tasks. Thus, we demonstrate the potential of learning from multiple data sources and modalities, underscoring the need to collect more holistic patient data that facilitates the application of multimodal ML in the healthcare domain. Our system is also provided as an open-source codebase to allow clinicians and researchers to train and test their own multimodal AI/ML systems more easily with local datasets, pre-trained feature extractors, and their own clinical questions. While our systematic evaluation of the effects of multiple data modality additions to our AI/ML framework was based on the MIMIC-IV dataset, this input was only used to exemplify our pipeline and to provide strong empirical evidence on the dynamics of performance from the use of different data modalities in a canonical HER scenario. The downstream trained models generated for this investigation could potentially be used in the future by people interested in predicting the demonstrated clinical tasks within intensive care units (ICUs) using multimodal data. However, we primarily encourage users to use our codebase to process their own EHR datasets and train predictive tasks of interest to them with the help of our pipeline. We envision a broad utility for the HAIM framework and its subprocesses focusing on driving cost-effective AI/ML activities for clinical and non-clinical operations. We hope that our HAIM framework can help reduce the time required to develop relevant AI/ML systems while efficiently utilizing human, financial, and digital resources in a more timely and unified approach than the current methods used in healthcare organizations.

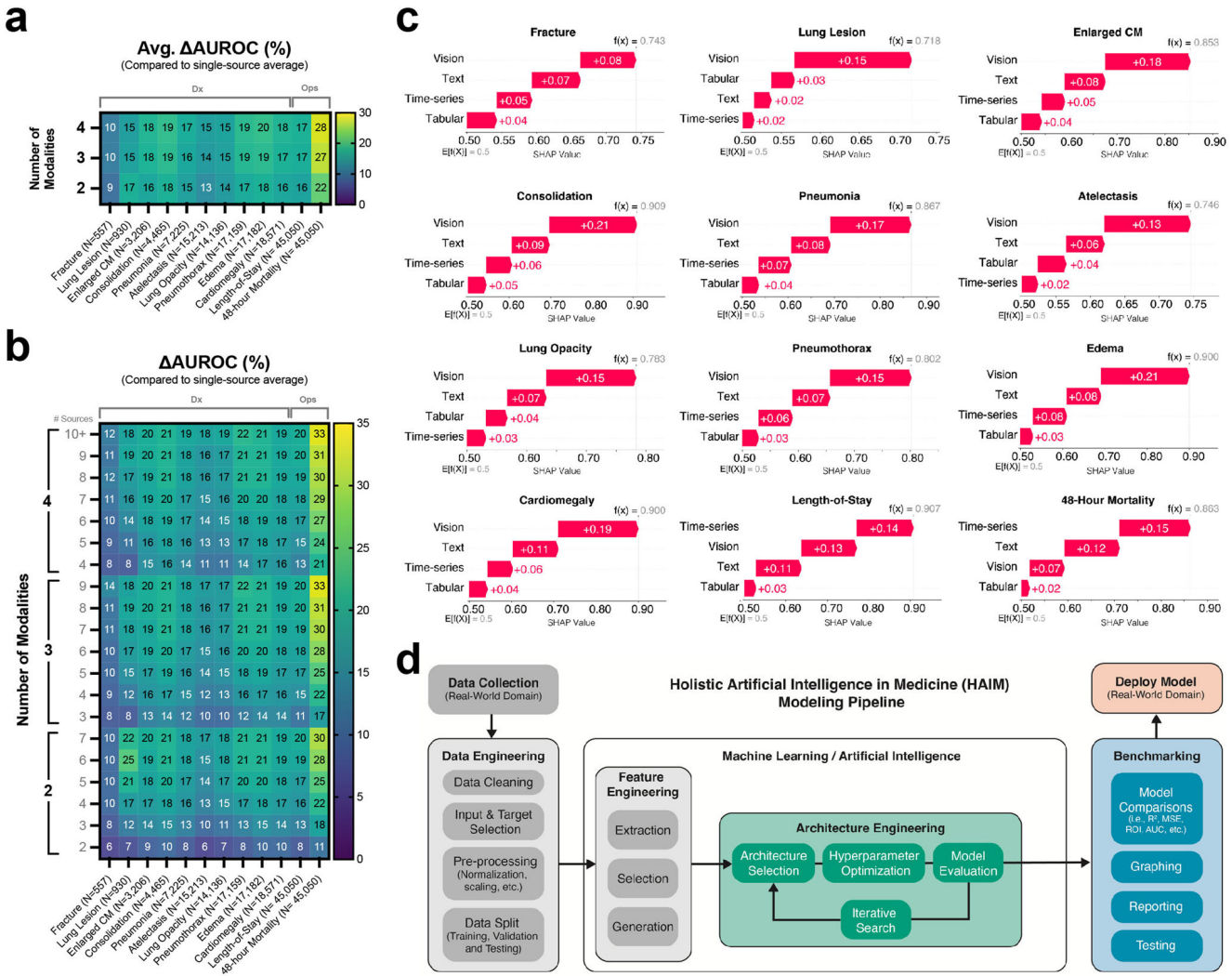

**Fig. 3 Multimodal HAIM framework is a flexible and robust method to improve predictive capacity for healthcare machine learning systems as compared to single-modality approaches. a** Average percent change of area under the receiver operating characteristic curve (Avg. ΔAUROC) for all tested multimodality HAIM models as compared to their single-source single-modality counterparts. While different models exhibit varying degrees of improvement, all tested models show positive Avg. ΔAUROC percentages. The number of modalities refers to the coverage among tabular, time-series, text, and image data. The number of sources refers to the coverage among available input data sources (10 for pathology diagnosis, 11 for length-of-stay, and 48 h mortality prediction). Thus, the position (Modality = 2, Sources = 3) corresponds to the average AUROC of all models across all input combinations covering any 2 modalities using any 3 input sources. **b** Expanded Avg. ΔAUROC percentages for all tested multimodality HAIM models and ordered by the number of used modalities (i.e., tabular, time-series, text, or images) as well as the number of used data sources. **c** Waterfall plots of aggregated Shapley values for independent data modalities per predictive task. While Shapley values for all data modalities appear to be positively contributing to the predictive capacity of all models, different tasks exhibit distinct distributions of aggregated Shapley values. **d** High-level schematic of the HAIM pipeline developed to support the presented work. After data collection or sourcing (HAIM-MIMIC-MM for this work), a process of feature selection and embedding extraction is applied to feed fusion embeddings into a process of iterative architecture engineering (model and hyperparameter selection). After particular models are selected and trained, they can be benchmarked to test and report results. This process concludes by the selection of a model for deployment in a use case scenario.

## METHODS

### Dataset

For this work, we utilize the Medical Information Cart for Intensive Care (MIMIC)-IV[32,33], an openly accessible database that contains de-identified records of 383,220 individual patients admitted to the ICU or emergency department (ED) of Beth Israel Deaconess Medical Center (BIDMC) in Boston, MA, USA, between 2008 and 2019 (inclusive). MIMIC-IV's most recent version (v1.0) improves on MIMIC-III[34] to provide public access to the EHR data of over 40,000 hospitalized patients based on the BIDMC's MetaVision clinical information system. We selected MIMIC-IV due to its large-scale, detailed documentation, generalizable formatting, corroborated use in AI/ML applications[35], and prior evaluations in terms of AI/ML interpretability, fairness, and bias[36]. To augment BIDMC's MIMIC-IV v1.0, we used the MIMIC Chest X-ray (CXR) database v2.0.0[37] containing 377,110 radiology images with free-text reports representing 227,835 medical imaging events that can be matched to corresponding patients included in MIMIC-IV v1.0. Both databases have been independently de-identified by deleting all personal health information, following the US Health Insurance Portability and Accountability Act of 1996 Safe Harbor requirements. After getting credentialled access from PhysioNet, we combined MIMIC-IV v1.0 and MIMIC-CXR-JPG v2.0.0 into a unified multimodal dataset (HAIM-MIMIC-MM) based on matched patient, admission, and imaging-study identifiers (i.e., subject_id, stay_id, study_id from MIMIC-IV and MIMIC-CXR-JPG databases). We used HAIM-MIMIC-MM throughout this study to test all the presented ML use cases analyzing various combinations of structured patient information, time-series data, medical images, and unstructured text notes, as presented in the following sections.

### Patient-centric data representation

We generated the individual files containing patient-specific information for single hospital admissions by querying the aggregated multimodal dataset HAIM-MIMIC-MM. Every HAIM-EHR file contains the details of current and previous patient admissions, transfers, demographics, laboratory measurements, provider orders, microbiology cultures, medication administrations, prescriptions, procedure events, intravenous and fluid inputs, sensor outputs, measurement events, radiological images, radiological reports, electrocardiogram reports, echocardiogram reports, notes, hospital billing information (e.g., diagnosis and procedure-related codes), as well as other time-stamped and charted information. The samples, therefore, include all available patient data collected within a specific admission and stay with all prior information occurring before the discharge or death time stamp. We stored all the individual patient files in HAIM-MIMIC-MM as "pickle" python-language object structures for ease of processing in subsequent sampling and modeling tasks. The code to generate the aggregated HAIM-MIMIC-MM dataset from credentialled access to MIMIC-IV v1.0 and MIMIC-CXR-JPG v2.0.0 datasets is available at our PhysioNet repository (https://doi.org/10.13026/dxcx-n572)[38] as well as our GitHub repository (https://github.com/lrsoenksen/HAIM). In addition, samples of pre-processed pickle patient files of HAIM-MIMIC-MM can be found in our PhysioNet project page https://doi.org/10.13026/dxcx-n572)[38]. A schematic of this patient-centric data representation as multimodal input for our HAIM framework is shown in Fig. 1.

### Patient data processing and multimodal feature extraction

We processed each HAIM-EHR patient file individually to generate fixed-dimensional vector embeddings for each of the possible input types, including all patient information from the time of admission until the selected inference event (e.g., time of imaging procedure for pathology diagnosis or end-of-day for 48 h mortality predictions). The generated embeddings from input modalities include: tabular data such as demographics ($E_{de}$ = demographics), structured time-series events ($E_{ce}$ = chart events, $E_{le}$ = laboratory events, $E_{pe}$ = procedure events), unstructured free text ($E_{radn}$ = radiological notes, $E_{ecgn}$ = electrocardiogram notes, $E_{econ}$ = echocardiogram notes), single-image vision ($E_{vp}$ = visual probabilities, $E_{vd}$ = visual dense-layer features) and multi-image vision ($E_{vmp}$ = aggregated visual probabilities, $E_{vmd}$ = aggregated visual dense-layer features). From these, patient signals used as time-series for embedding extraction (classified by type of event) can be found in Supplementary Table 1. We then implemented fixed embedding extraction procedures based on standard data modalities (i.e., tabular data, time-series, text, and images) to reduce its dependence on site-specific data architectures and allow for a consistent embedding format that may be applied to arbitrary ML pipelines. Note that throughout this work, we refer to data "modality" as a distinct term to data "source", where the former is used to define broad classes of data usually digitalized in different format types, while the latter simply refers to different input variables belonging to a data modality as defined in Supplementary Table 2.

We extracted the embeddings based solely on tabulated demographics data ($E_{de}$) by querying normalized numerical values from the patient record. We obtained time-series embeddings using time-stamped data from the structured patient chart, laboratory, and procedure event lists (i.e., $E_c$ $E_{le}$, $E_{pe}$, respectively). We selected a set of key clinical signals for each type of event list and constructed the corresponding time sequences from the time of patient admission to the time-stamp allowable for each individual feature (see Supplementary Table 1). The embeddings encode the signal length, maximum, minimum, mean, median, SD, variance, number of peaks, and average time-series slope and piece-wise change over time of these metrics. The time-series signals for $E_{ce}$ include: heart rate (HR), non-invasive systolic blood pressure (NBP$_s$), non-invasive diastolic blood pressure (NBP$_d$), respiratory rate, oxygen saturation by pulse oximetry (SpO$_2$), Glasgow coma scales (GCS) for verbal, eye, and motor response (GCS$_V$, GCS$_E$, GCS$_M$ respectively). Moreover, time-series $E_{le}$ include: glucose, potassium, sodium, chloride, creatinine, urea nitrogen, bicarbonate, anion gap, hemoglobin, hematocrit, magnesium, platelet count, phosphate, white blood cells, total calcium, mean corpuscular hemoglobin (MCH), red blood cells, mean corpuscular hemoglobin concentration, mean corpuscular volume, red blood cell distribution width, platelet count, neutrophils, vancomycin. Lastly, time-series $E_{pe}$ procedures include: foley catheter, peripherally inserted central catheter (PICC), intubation, peritoneal dialysis, bronchoscopy, electroencephalogram (EEG), dialysis with continuous renal replacement therapy, dialysis with catheter, removed chest tubes, and hemodialysis.

We obtained embeddings for the unstructured free text ($E_{radn}$, $E_{ecgn}$, and $E_{econ}$) by concatenating all available text from each of these types of notes as continuous strings and then by processing them using Clinical BERT[39], a transformer-based bidirectional encoder model pre-trained on a large corpus of biomedical and medical text. This transformer-based model generates a single 768-dimensional vector, or embedding, per unstructured text type. We split notes longer than the maximum input token size for Clinical BERT (i.e., 512 tokens) into the smallest number of processable text chunks to generate various embeddings sequentially, all of which are averaged to produce a single 768-dimensional output embedding for the entire text.

Finally, we processed vision data included in this work using a pre-trained Densenet121 convolutional neural network (CNN) previously fine-tuned on the X-ray CheXpert dataset[40] (i.e., Densenet121-res224-chex)[41]. We selected this model because the availability of at least one time-stamped chest X-ray per patient file within the HAIM-MIMIC-MM database as its core visual component. Densenet121-res224-chex is part of TorchXRayVision, a unified library, and repository of datasets and SOTA pre-trained models for chest pathology classification using X-rays[41]. While other computer vision models pre-trained on large sets of medical imaging data may be utilized to extract embeddings within the HAIM framework, for the purpose of experimentally validating our pipeline, we used Densenet121-res224-chex as a canonical method to extract visual embeddings. We obtained the single-image embeddings per HAIM-EHR patient file by rescaling each image into $224 \times 224$ size using a standard interpolation method with resampling using pixel area relations, and then feeding it into the selected network to extract: (a) output class probabilities and (b) final dense-layer features. The output classes per image are the 18-dimensional diagnosis probability vector generated directly by Densenet121-res224-chex, which produces the embedding $E_{vp}$. The dense network features per image are the 1024-dimensional vector generated by extracting the outputs of the last dense layer of the model, which produces the embedding $E_{vd}$. Multi-image embeddings are also obtained by averaging feature-wise the output class probabilities and dense-feature embeddings of all available images per HAIM-EHR patient file (e.g., X-ray studies with multiple planes and past X-ray studies). This produces an aggregated multi-image diagnosis probability embedding ($E_{vmp}$) and multi-image dense-layer embedding ($E_{vmd}$) per patient that considers all available X-rays and not only the most recent one.

There are various advantages of using SOTA pre-trained models specific to each data modality (i.e., tabular, time-series, text, and images) such as Clinical BERT[39] and Densenet121-res224-chex[41] as feature extractors in our HAIM framework. First, every single-date pre-trained SOTA model can be user-defined and easily exchanged with updated ones, as long as their respective dense features or embeddings are accessible. This departs from other multimodal AI/ML strategies that attempt to directly fuse heterogeneous input data, which makes these systems less modular and usually incompatible with the use of high-performing open-source single-data-type models produced by other organizations and researchers[10,29]. A second advantage of using SOTA feature extractors within our framework is that users can easily generate unified input vectors to focus primarily on downstream modeling and rapid training of their predictive systems of interest, which can accelerate deployment.

In our sample demonstration of the HAIM framework using the HAIM-MIMIC-MM database, the dimensionality of each of these embeddings is $E_{de} = 6$, $E_{ce} = 99$, $E_{le} = 242$, $E_{pe} = 110$, $E_{radn} = 768$, $E_{ecgn} = 768$, $E_{econ} = 768$, $E_{vp} = 18$, $E_{vd} = 18$, $E_{vmp} = 1024$, and $E_{vmd} = 1024$. Detail on the presence and handling of missing input data is provided as part of Supplementary Table 3. Once all single-modality embeddings are generated, we flatten, normalize, and concatenate them into a single one-dimensional multimodal fusion embedding per HAIM-EHR patient file, which constitutes the input for all downstream modeling tasks in our HAIM framework (see Supplementary Fig. 3 for algorithmic detail of such process). This deep patient representation in vector form can be made of fixed size within or across healthcare institutions (4845-dimensional for this work), which can allow for rapid iteration in the development of generic ML systems for relevant predictive analytics in various applications.

## Modeling
After we extracted all multimodal fusion embeddings for all HAIM-EHR patient files in the HAIM-MIMIC-MM database, we generated classification models across various clinical and operational tasks, including: (a) chest pathology diagnosis, (b) length-of-stay and (c) 48 h mortality predictions. For each of these modeling tasks, we split the available embeddings randomly into training (80%) and testing (20%) sets 5 times (with 5 different splits), stratifying by patients during our experiments to avoid data leakage of patient-level information from training to testing, compute SDs, and to ensure adequate comparison of recorded predictive values. For the chest pathology diagnosis tasks, we applied an additional stratification by pathology to balance the target ratios. We then conducted experiments to compare the effect of all different combinations of input data modalities and sources using the extracted multimodal fusion embeddings as presented in further sections. An algorithmic formulation of our HAIM framework in the context of the data processing, feature extraction, and downstream predictive task modeling stages is provided as part of Supplementary Fig. 3. Detail on the sensitivity of missing input data to downstream predictions is also provided as part of Supplementary Fig. 4.

## Tasks of interest
*Chest pathology diagnosis prediction.* Early detection of certain pathologies in CT scans and other diagnostic imaging modalities enables clinicians to focus on early intervention rather than delayed treatment for advanced stages of relevant pathologies. Within this task of interest, we chose to target the prediction of 10 common thorax-level pathologies (i.e., fractures, lung lesions, enlarged cardio mediastinum, consolidation, pneumonia, lung opacities, atelectasis, pneumothorax, edema, and cardiomegaly) that can be typically assessed by radiologists through chest X-ray, to demonstrate that HAIM outperforms image-only approaches. The ground-truth values for each chest pathology included in HAIM-MIMIC-MM were derived from MIMIC-CXR-JPG v2.0.0, where radiology notes were processed to determine if each of these pathologies was explicitly confirmed as present (value = 1), explicitly confirmed as absent (value = 0), inconclusive in the study (value = −1), or not explored (no value). We only selected samples with 0 or 1 values, removing the rest from the training and testing data. Thus, for this specific task, we utilized the multimodal fusion embeddings as input and the ground-truth chest pathology HAIM-MIMIC-MM values as the output target to predict. From these embeddings, we only excluded the unstructured radiology notes component ($E_{rad}$) from the allowable input to avoid potential overfitting or misrepresentations of real predictive value. We trained and tested independent binary classification models for each target chest pathology and input source combination as described in the general model training setup section. Final sample sizes for each pathology diagnosis task are: Fracture ($N = 557$), Lung Lesion ($N = 930$), Enlarged cardio mediastinum ($N = 3206$), Consolidation ($N = 4465$), Pneumonia ($N = 7225$), Lung opacity ($N = 14,136$), Atelectasis ($N = 15,213$), Pneumothorax ($N = 17,159$), Edema ($N = 17,182$) and Cardiomegaly ($N = 18,571$).

*Length-of-Stay prediction.* Projected patient length-of-stay plays a vital role for both patients and hospital systems in making informed medical and economic decisions. An accurate forecast of patient stay enhances patient satisfaction, hospital resource allocations, and doctors' ability to make more effective treatment planning[42]. Particularly, predicting next 48 h discharges is critical for physicians to identify and prioritize patients ready for discharge and for case management teams to accelerate discharge preparations, which ultimately reduces patient burden and direct operating costs in healthcare systems[43]. To demonstrate the HAIM framework for healthcare operations tasks, we predicted whether or not a patient will be discharged without expiration during the next 48 h as a binary classification problem: discharged alive ≤48 h (1) or otherwise (0). In case of patient death, we set the class label to 0. Each sample in this predictive task corresponds to a single patient-admission EHR time point where an X-ray image was obtained ($N = 45{,}050$).

*48 h mortality prediction.* Due to its time and outcome-critical environments, clinicians in ICU units often need to make rapid evaluations of patient conditions to inform treatment plans[44]. However, current standards of estimating patient severity, such as the Acute Physiologic Assessment and Chronic Health Evaluation score, fail to incorporate medical characteristics beyond acute physiology[45]. Accurate mortality prediction can give clinicians advanced warnings of possible deteriorations and share the burdens of making information-heavy decisions[44]. To further demonstrate the versatility of the HAIM framework, we also built models to predict the probability that a patient will expire during the next 48 h as a binary classification problem: expired ≤48 h (1) or otherwise (0). In the case of a patient whose hospital exit status is not expiration, we set the class label to 0. It should be noted that a patient can acquire different target class labels at different time points during their stay due to changes in status and proximity to the discharge or time of death. Similar to the length-of-stay modeling, each sample in this predictive task corresponds to a single patient-admission EHR time point where an X-ray image was obtained ($N = 45{,}050$).

## General model training setup

We initially explored seven ML architectures, including logistic regression, classification and regression trees, random forest, multi-layer perceptron, gradient boosted trees (XGBoost), gradient boosting machines (LightGBM), as well as attentive tabular networks TabNet to heuristically decide on the best model choice for follow-up experiments. Since XGBoost supports fast computations for large-scale experiments and consistently outperformed other architectures during preliminary observations, we selected this canonical methodology for all further tests. Our XGBoost-based modeling experiments were conducted using every possible combination of input embeddings, extracted as described in previous sections, from the allowable 11 data sources (i.e., $E_{de}$, $E_{ce}$, $E_{pe}$, $E_{le}$, $E_{ecgn}$, $E_{econ}$, $E_{radn}$, $E_{vp}$, $E_{vd}$, $E_{vmp}$, and $E_{vmd}$) and 4 modalities (i.e., tabular, time-series, text, and images). In this process, we concatenated each data stream permutation to produce fusion embeddings and train XGBoost models using single-modality ($N_{1M} = 52$), double-modality ($N_{2M} = 392$), triple-modality ($N_{3M} = 972$) and quadruple-modality ($N_{4M} = 630$) combination of inputs. This corresponds to the generation of 2047 models (per predictive task) for the cases of length-of-stay and 48 h mortality. As previously mentioned, in the case of chest pathology diagnosis, the embeddings corresponding to all radiology notes ($E_{radn}$) are not included as part of the input fusion embeddings to allow for fair comparison with the output target, which was originally determined from examining notes in MIMIC-CXR-JPG. This reduced the total number of possible models per chest pathology diagnosis task to 1023 ($N_{1M} = 26$, $N_{2M} = 196$, $N_{3M} = 486$, $N_{4M} = 315$). Since there are ten chest pathologies, defined as binary classification problems for our experiments, we trained a total of $1023*10 = 10{,}230$ models for chest pathology diagnosis prediction. As mentioned previously, all XGBoost models were trained five times with five different data splits to repeat the experiments and compute average metrics and SDs.

All defined models ($N_{Models} = 14{,}324$) were trained and tested to evaluate the advantage of multimodal predictive systems, based on the HAIM framework, as compared to single modality ones for the aforementioned clinical and operational tasks. We capture average trends of model performance by reporting the average area under the receiver operating characteristic (AUROC) curve on the testing set (20%) over five consecutive iterations of randomized train-test data splitting and model training. The hyperparameter combinations of individual XGBoost models were selected within each training loop using a fivefold cross-validated grid search on the training set (80%). This XGBoost tuning process selected the maximum depth of the trees (5–8), the number of estimators (200 or 300), and the learning rate (0.05, 0.1, 0.3) according to the parameter value combination leading to the highest observed AUROC within the training loop. This model cross-validation strategy at the level of each data source combination ensures that the respective test sets are never used for model training, model selection, model comparison, or reporting across any of the 14,324 uniquely trained models. Thus, throughout this study, the test set remains unseen at the level of each model for all models, which minimizes the potential for data leakage or model selection overfitting.

The aggregated test set performance metrics (fivefold test averages and SDs) of all these models grouped by the number of data sources and modalities can be found in Fig. 2. We conducted all embedding generation and computational experiments using a parallelization strategy under MIT's Supercloud server (https://supercloud.mit.edu) with 30GB RAM and 1 NVIDIA Tesla V100 Volta graphics processing unit per instance. A high-level schematic representation of the HAIM framework, from data sourcing to model benchmarking, can be found in Fig. 3.

## REFERENCES

1. Topol, E. *Deep medicine: how artificial intelligence can make healthcare human again.* (Hachette UK, 2019).
2. Huang, S.-C., Pareek, A., Seyyedi, S., Banerjee, I. & Lungren, M. P. Fusion of medical imaging and electronic health records using deep learning: a systematic review and implementation guidelines. *NPJ Dig. Med.* **3**, 1–9 (2020).

3. Gietzelt, M., Löpprich, M., Karmen, C. & Ganzinger, M. Models and data sources used in systems medicine. *Methods Inf. Med.* **55**, 107–113 (2016).

4. Boonn, W. W. & Langlotz, C. P. Radiologist use of and perceived need for patient data access. *J. Dig. imaging* **22**, 357–362 (2009).

5. Wang, W. & Krishnan, E. Big data and clinicians: a review on the state of the science. *JMIR Med. Inform.* **2**, e1 (2014).

6. Sun, W. et al. Data processing and text mining technologies on electronic medical records: a review.*J. Healthcare Eng.* **2018**, 4–7 (2018).

7. Agrawal, S. et al. Selection of 51 predictors from 13,782 candidate multimodal features using machine learning improves coronary artery disease prediction. *Patterns* **2**, 100364 (2021).

8. Bagheri, A. et al. Multimodal learning for cardiovascular risk prediction using EHR data. *arXiv preprint arXiv:2008.11979* (2020).

9. Li, P., Hu, Y. & Liu, Z.-P. Prediction of cardiovascular diseases by integrating multimodal features with machine learning methods. *Biomed. Signal Process. Control* **66**, 102474 (2021).

10. Liu, Y. et al. A deep learning system for differential diagnosis of skin diseases. *Nat. Med.* **26**, 900–908 (2020).

11. Stidham, R. W. Artificial Intelligence for Understanding Imaging, Text, and Data in Gastroenterology. *Gastroenterol. Hepatol.* **16**, 341 (2020).

12. Paquette, A. G., Hood, L., Price, N. D. & Sadovsky, Y. Deep Phenotyping During Pregnancy for Delivery of Predictive and Preventive Medicine. *Sci.Transl. Med.* **12**, 2–4 (2020).

13. Purwar, S., Tripathi, R. K., Ranjan, R. & Saxena, R. Detection of microcytic hypochromia using cbc and blood film features extracted from convolution neural network by different classifiers. *Multimed. Tools Appl.* **79**, 4573–4595 (2020).

14. Hügle, M., Kalweit, G., Hügle, T. & Boedecker, J. In *Explainable AI in Healthcare and Medicine* 79–92 (Springer, 2021).

15. Tomašev, N. et al. A clinically applicable approach to continuous prediction of future acute kidney injury. *Nature* **572**, 116–119 (2019).

16. Ieracitano, C., Mammone, N., Hussain, A. & Morabito, F. C. A novel multi-modal machine learning based approach for automatic classification of EEG recordings in dementia. *Neural Netw.* **123**, 176–190 (2020).

17. Prashanth, R., Roy, S. D., Mandal, P. K. & Ghosh, S. High-accuracy detection of early Parkinson's disease through multimodal features and machine learning. *Int. J. Med. Inform.* **90**, 13–21 (2016).

18. Hyun, S. H., Ahn, M. S., Koh, Y. W. & Lee, S. J. A machine-learning approach using PET-based radiomics to predict the histological subtypes of lung cancer. *Clin. Nucl. Med.* **44**, 956–960 (2019).

19. Yala, A., Lehman, C., Schuster, T., Portnoi, T. & Barzilay, R. A deep learning mammography-based model for improved breast cancer risk prediction. *Radiology* **292**, 60–66 (2019).

20. Reda, I. et al. Deep learning role in early diagnosis of prostate cancer. *Technol. Cancer Res. Treat.* **17**, 1533034618775530 (2018).

21. An, G. et al. Comparison of machine-learning classification models for glaucoma management. *J. Healthcare Eng.* **2018**, 2–7 (2018).

22. Patel, M. J. et al. Machine learning approaches for integrating clinical and imaging features in late-life depression classification and response prediction. *Int. J. Geriatr. Psychiatry* **30**, 1056–1067 (2015).

23. Huang, S.-C., Pareek, A., Zamanian, R., Banerjee, I. & Lungren, M. P. Multimodal fusion with deep neural networks for leveraging CT imaging and electronic health record: a case-study in pulmonary embolism detection. *Sci. Rep.* **10**, 1–9 (2020).

24. Tiulpin, A. et al. Multimodal machine learning-based knee osteoarthritis progression prediction from plain radiographs and clinical data. *Sci. Rep.* **9**, 1–11 (2019).

25. Wu, J. et al. Radiological tumour classification across imaging modality and histology. *Nat. Mach. Intell.* **3**, 787–798 (2021).

26. Mei, X. et al. Artificial intelligence–enabled rapid diagnosis of patients with COVID-19. *Nat. Med.* **26**, 1224–1228 (2020).

27. Bardak, B. & Tan, M. Improving clinical outcome predictions using convolution over medical entities with multimodal learning. *Artif. Intell. Med.* **117**, 102112 (2021).

28. Jin, M. et al. Improving hospital mortality prediction with medical named entities and multimodal learning. *arXiv preprint arXiv:1811.12276* (2018).

29. Rajkomar, A. et al. Scalable and accurate deep learning with electronic health records. *NPJ Digital Med.* **1**, 1–10 (2018).

30. Li, Y. et al. Inferring multimodal latent topics from electronic health records. *Nat. Commun.* **11**, 1–17 (2020).

31. Štrumbelj, E. & Kononenko, I. Explaining prediction models and individual predictions with feature contributions. *Knowl. Inf. Syst.* **41**, 647–665 (2014).

32. Johnson, A. et al. MIMIC-IV (version 1.0). *PhysioNet,* https://doi.org/10.13026/s6n6-xd98. (2021).

33. Goldberger, A. L. et al. PhysioBank, PhysioToolkit, and PhysioNet: components of a new research resource for complex physiologic signals. *Circulation* **101**, e215–e220 (2000).

34. Johnson, A. E. et al. MIMIC-III, a freely accessible critical care database. *Sci. Data* **3**, 1–9 (2016).

35. Royalty, J. P. Machine Learning Time-to-Event Mortality Prediction in MIMIC-IV Critical Care Database (Doctoral dissertation). *Undergraduate Research Scholars Program.* Available electronically from https://hdl.handle.net/1969.1/194429 (2021).

36. Meng, C., Trinh, L., Xu, N. & Liu, Y. MIMIC-IF: Interpretability and Fairness Evaluation of Deep Learning Models on MIMIC-IV Dataset. *arXiv preprint arXiv:2102.06761* (2021).

37. Johnson, A. E. et al. MIMIC-CXR-JPG, a large publicly available database of labeled chest radiographs. *arXiv preprint arXiv:1901.07042* (2019).

38. Soenksen, L. R. & Ma, Y. Code for generating the HAIM multimodal dataset of MIMIC-IV clinical data and x-rays (version 1.0.0). *PhysioNet,* https://doi.org/10.13026/dxcx-n572 (2022).

39. Alsentzer, E. et al. Publicly available clinical BERT embeddings. *arXiv preprint arXiv:1904.03323* (2019).

40. Irvin, J. et al. Chexpert: a large chest radiograph dataset with uncertainty labels and expert comparison. *Proc. AAAI Conf. Artif. Intell.* **33**, 590–597 (2019).

41. Cohen, J. P. et al. TorchXRayVision: A library of chest X-ray datasets and models. *arXiv preprint arXiv:2111.00595* (2021).

42. Bertsimas, D., Pauphilet, J., Stevens, J. & Tandon, M. Predicting inpatient flow at a major hospital using interpretable analytics. *Manufact. Service Operations Manag.* **1**, 1–4 (2021).

43. Zhu, T., Luo, L., Zhang, X., Shi, Y. & Shen, W. Time-series approaches for forecasting the number of hospital daily discharged inpatients. *IEEE J. Biomed. Health Inform.* **21**, 515–526 (2015).

44. Awad, A., Bader-El-Den, M., McNicholas, J. & Briggs, J. Early hospital mortality prediction of intensive care unit patients using an ensemble learning approach. *Int. J. Med. Inform.* **108**, 185–195 (2017).

45. Awad, A., Bader–El–Den, M. & McNicholas, J. Patient length of stay and mortality prediction: a survey. *Health Serv. Manag. Res.* **30**, 105–120 (2017).

## ACKNOWLEDGEMENTS

We thank the PhysioNet team from the MIT Laboratory for Computational Physiology for providing our researchers with credentialed access to the MIMIC-IV and MIMIC-CXR-JPG datasets and for their support in guiding multimodal data interrogation and consolidation. We especially thank Leo A. Celi and Sicheng Hao for their support on MIMIC-IV data review, as well as the Harvard TH Chan School of Public Health, Harvard Medical School, the Institute for Medical Engineering and Science at MIT, and the Beth Israel Deaconess Medical Centre for their continued support of this work. We thank the MIT Supercloud for their support and help in setting up a workspace as well as offering technical advice throughout the project. Finally, we thank Eli Pivo for providing feedback and support on computational experiments to our work. This work was supported by the Abdul Latif Jameel Clinic for Machine Learning in Health (L.R.S., D.B., and I.F). H.W. is supported by the National Science Foundation Graduate Research Fellowship under Grant No. 174530. Any opinion, findings, conclusions, or recommendations expressed in this material are those of the authors and do not necessarily reflect the views of the National Science Foundation.

## AUTHOR CONTRIBUTIONS

L.R.S., Y.M., C.Z., L.B. planned and performed experiments, wrote code, analyzed the data, and wrote the paper. K.V.C., L.N., H.M.W., M.L.L. performed experiments, wrote code, analyzed the data, and edited the paper. I.F. contributed to research design and edited the paper. D.B. directed overall research and edited the paper. In aggregate, L.R.S., Y.M., C.Z., L.B., K.V.C., L.N. contributed equally to this work.

## COMPETING INTERESTS
The authors declare no competing interests.

## ETHICS APPROVAL:
*Human subject research:* This work only makes use of MIMIC-IV v1.0[32] and MIMIC Chest X-ray (CXR) v2.0.0[37] to generate the multimodal HAIM-MIMIC-MM dataset and does not contain any additional information involving human participants obtained by the authors.

## ADDITIONAL INFORMATION

**Supplementary information** The online version contains supplementary material available at https://doi.org/10.1038/s41746-022-00689-4.

**Correspondence** and requests for materials should be addressed to Dimitris Bertsimas.

**Reprints and permission information** is available at http://www.nature.com/reprints

**Publisher's note** Springer Nature remains neutral with regard to jurisdictional claims in published maps and institutional affiliations.

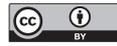

**Open Access** This article is licensed under a Creative Commons Attribution 4.0 International License, which permits use, sharing, adaptation, distribution and reproduction in any medium or format, as long as you give appropriate credit to the original author(s) and the source, provide a link to the Creative Commons license, and indicate if changes were made. The images or other third party material in this article are included in the article's Creative Commons license, unless indicated otherwise in a credit line to the material. If material is not included in the article's Creative Commons license and your intended use is not permitted by statutory regulation or exceeds the permitted use, you will need to obtain permission directly from the copyright holder. To view a copy of this license, visit http://creativecommons.org/licenses/by/4.0/.